\documentclass{article}

\usepackage{graphicx,array}
\usepackage{graphicx}
\usepackage{color}
\usepackage{subcaption}
\usepackage{xargs}   
\usepackage[colorinlistoftodos,prependcaption,textsize=tiny]{todonotes}
\newcommandx{\change}[2][1=]{\todo[linecolor=blue,backgroundcolor=blue!25,bordercolor=blue,#1]{#2}}
\newcommandx{\info}[2][1=]{\todo[linecolor=OliveGreen,backgroundcolor=OliveGreen!25,bordercolor=OliveGreen,#1]{#2}}
\newcommandx{\improvement}[2][1=]{\todo[linecolor=red,backgroundcolor=red!25,bordercolor=red,#1]{#2}}

\usepackage{arxiv}

\usepackage[utf8]{inputenc} 
\usepackage[T1]{fontenc}    
\usepackage{hyperref}       
\usepackage{url}            
\usepackage{booktabs}       
\usepackage{amsfonts}       
\usepackage{nicefrac}       
\usepackage{microtype}      
\usepackage{lipsum}

\title{Weight Update Skipping: Reducing Training Time for Artificial Neural Networks}

\author{
  Pooneh Safayenikoo \\
  Electrical Engineering and Computer Science Department\\
  University of Missouri, Columbia\\
  \texttt{safayenikoop@mail.missouri.edu} \\
  \And
  Ismail Akturk \\  
  Electrical Engineering and Computer Science Department\\
  University of Missouri, Columbia\\
  \texttt{akturki@missouri.edu}
}
\begin{document}

\maketitle

\begin{abstract}
Artificial Neural Networks (ANNs) are known as state-of-the-art techniques in Machine Learning (ML) and have achieved outstanding results in data-intensive applications, such as recognition, classification, and segmentation. These networks mostly use deep layers of convolution or fully connected layers with many filters in each layer, demanding a large amount of data and tunable hyperparameters to achieve competitive accuracy. As a result, storage, communication, and computational costs of training (in particular training time) become limiting factors to scale them up. In this paper, we propose a new training methodology for ANNs that exploits the observation of 
improvement of accuracy
shows temporal variations which allows us to skip updating weights when the variation is minuscule. During such time windows, we keep updating bias which ensures the network still trains and avoids overfitting; however, we selectively skip updating weights (and their time consuming computations). Such a training approach virtually achieves the same accuracy with considerably less computational cost, thus lower training time. We propose two methods for updating weights and evaluate them by analyzing four state-of-the-art models, AlexNet, VGG-11, VGG-16, ResNet-18 on CIFAR datasets. On average, our two proposed methods called WUS and WUS+LR reduced the training time (compared to the baseline) by 54\%, and 50\%, respectively on CIFAR-10; and 43\% and 35\% on CIFAR-100, respectively.
\end{abstract}


\section{Introduction}
Artificial Neural Networks (ANNs) have recently gained a lot of attention to solve highly non-linear problems like recognition, classification, and segmentation. The solution mostly obtained using a network of deep convolutional and/or fully connected layers with many filters in each layer~\cite{zurada1992introduction},~\cite{ han2016eie}. As the size of the network grows, the size of the dataset becomes a key factor in the performance of the trained network. The larger the dataset is, the less likely the overfitting happens, assuming sufficient variance exists in the dataset. Training ANNs using a large dataset is computationally expensive if one uses online learning. On the other hand, batch learning requires a massive memory which often is not feasible for today's popular datasets. The minibatch learning provides a middle ground to the problem; however, the final accuracy and convergence time highly depend on properly shuffling the data when reading data into the mini-batch. Therefore, new challenges are introduced with larger (and deeper) neural networks: an extensive training time, computational cost, and overfitting.

Training time is one of the biggest concerns for today's ANNs, particularly for Deep Neural Networks (DNNs). As the depth of DNNs increases, the number of parameters to be updated also increases. As a result training takes longer.  Although modern computer systems with several GPUs are used for training, the converging for deep models such as ResNet~\cite{he2016deep}, VGG~\cite{simonyan2014very} and DenseNet~\cite{huang2016deep} can still take multiple weeks on a big dataset (e.g., ImageNet dataset~\cite{krizhevsky2012imagenet}).

The computational cost and memory bandwidth demands for training are typically an order of magnitude higher than inference. Yet, the overhead of training is increased by the size of data and the depth of neural networks. Thus, training DNN models on a large dataset is still a big concern in machine learning. There exist a great number of efforts to reduce the training overhead in DNNs, such as PruneTrain~\cite{lym2019prunetrain}, which is a cost-efficient technique that uses pruning to reduce the computation cost of the training. However, pruning may have its own side effects, including difficulties in generalization when the network is trained.
The modern DNNs learn complex patterns by updating several millions of parameters by examining the training data. While this allows the models to fit in the training dataset, its generalization ability may reduce for unseen data, known as overfitting. Many methods have been proposed to address the overfitting problem, such as regularization, data augmentation, and Dropout~\cite{hinton2012improving}.

On the other hand, shallower neural networks can be trained efficiently within a resealable time without overfitting problems. However, they may not capture complex patterns in the dataset which are essential to applications dealing with large unseen dataset. Therefore, there is a dilemma between shallower and deeper neural networks, and many challenging trade-offs in the landscape of machine learning, such as overfitting, generalization, training time, dataset size, number of parameters, and computational cost.

In this paper, we propose two new methods for training that try to find a balance among the above-mentioned trade-offs without degrading the accuracy of the NNs. Our approaches are to keep the general architecture of a deep neural as is (i.e., without pruning); however, we selectively choose the parameters to be updated during training based on improvement of accuracy which shows temporal variation. Proposed selective update approaches greatly reduce the time spent on training without reducing the accuracy and without inversely impacting the generalization property of the NNs.
We perform experiments on four state-of-the-art convolutional networks: AlexNet, VGG-11, and VGG-16, and ResNet-18 with  CIFAR datasets.  The results show that on average, our two proposed methods, called WUS and WUS+LR, can reduce the training time by 54\%, and 50\%, with an average accuracy loss of 0.71\% and 0.59\% respectively on CIFAR-10; and 43\% and 35\% on CIFAR-100 with on average accuracy loss of 1.05\% and 0.81\%, respectively. The details of WUS and WUS+LR are discussed in Section~\ref{sec:wus}. 

\section{Background and Motivation}
Numerous works have been proposed for training improvement of deep neural networks. One of the earlier works is called early stopping~\cite{prechelt1998early} that attempts to avoid the overfitting by stopping the training when the validation error starts to increase. 
Some other popular methods use this idea that all units in a deep network do not require to participate in the training step at each epoch. Dropout~\cite{hinton2012improving} randomly removes hidden units by multiplying each unit by Bernoulli random variable. Therefore, dropout reduces overfitting by making neural networks shallower. In contrast to these earlier works, our proposed method keeps the network structure intact, but reduces the number of weight updates (and eliminates the associated gradient calculation) in backpropagation when the accuracy improvement seems to be flattened. In particular, training does not end as opposed to early stopping, rather our method exploits the bias update to keep learning continue, but minimizes the number of parameter update during the time period when the accuracy seems to be stagnated. In a sense, our proposed method dynamically adjusts the degree of efforts need to be put on weight update, so minimizes the computational burden when it does not pay off, thus reduces the overall training time while fulfilling accuracy constraints.

Stochastic Depth~\cite{huang2016deep} uses the dropout idea (omitting the hidden nodes), however, it makes the network shallower by removing entire layers at a time with the identity function to reduce training time. FreezeOut~\cite{brock2017freezeout} also tries to reduce the training time by decreasing the learning rate gradually to zero (i.e., ending training). It maintains a separate learning rate for each layer, and a it reduces the learning rate starting from the first layer and moving to subsequent layers.  A layer whose learning rate reaches down to zero becomes frozen (i.e., no longer parameter update). 
They can achieve a 20\% speed-up with a 3\% accuracy loss. In contrast, our approach does not maintain separate learning rate for each layer; and it does not keep layers frozen indefinitely; rather it selectively allow (and disallow) parameter update for layers temporarily. Our approach can achieve around 50\% speed-up with less than 1\% accuracy loss. 


Batch Normalization~\cite{ioffe2015batch} is a method for significantly accelerating the training by normalizing the mean and variance of each hidden layer corresponding to each mini-batch. PruneTrain~\cite{lym2019prunetrain} uses three optimization technique: lasso regularization, channel union, and dynamic mini-batch adjustment to accelerate training. They start pruning from training to reduce the cost of computation and memory access. Eyeriss~\cite{chen2016eyeriss} uses a row-stationary dataflow on a spatial architecture to reduce the energy consumption of the convolutional neural network. All these efforts have their own merit and are orthogonal to our approach.

\begin{figure}
\centering
\includegraphics[height=10cm]{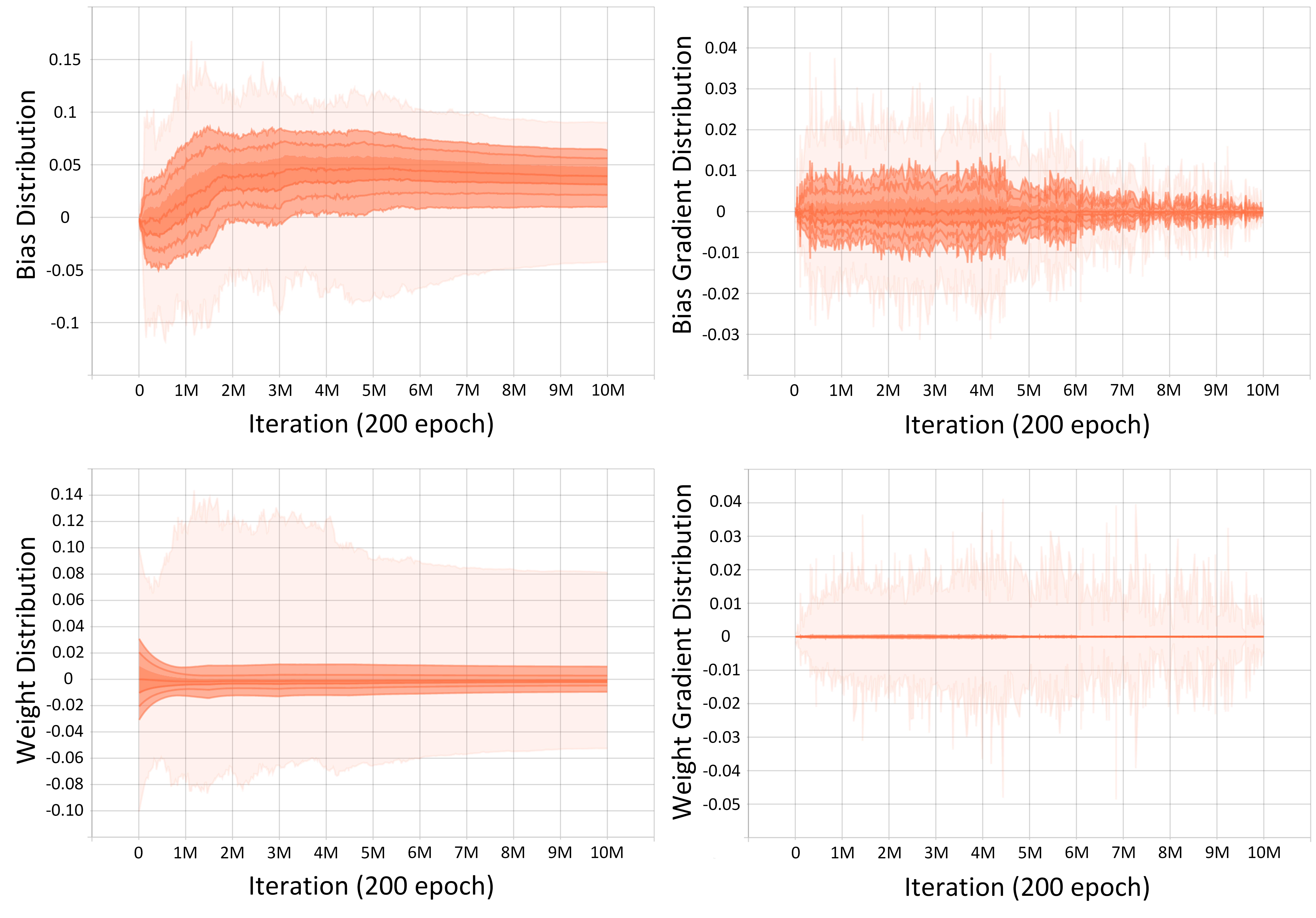}
\caption{The value distribution of weights and biases during training of the last convolution layer of VGG-11 for CIFAR-100 dataset.}
\label{fig-weight-bias-distribution}
\end{figure}

The goal of this paper is to reduce training time and pressure on computational resources available on the system. The intuition behind the proposed selective weight update methods is as follows. We observe that the values of weights tend to remain almost the same once the training reaches to certain point in training. Fig.~\ref{fig-weight-bias-distribution} 
shows the distribution of weight and bias values and their gradients for the eighth layer (that has 2359296 weights and 512 biases) of the VGG-11 during 200 epochs of training. In Fig.~\ref{fig-weight-bias-distribution}, the values of weights remain almost the same after 15 epochs. On the other hand, looking at bias values, one can observe that they keep changing for longer epochs. Therefore, we find out that by updating the bias alone for the duration of epochs in which the value of weights are changing at minuscule scale (compared to earlier epochs), one can achieve the almost the same accuracy. The intuition behind this is that the convergence for weights has happened sooner than bias. Therefore, by skipping weight updates (and their associated gradient calculations) and updating bias alone, we can reduce the training time without sacrificing accuracy, and without inversely impacting generalization of the NN (as the number of trainable parameters remain intact).

\vspace{-0.4cm}

\section{Weight Update Skipping (WUS)}
\label{sec:wus}
The proposed method, called Weight Update Skipping (WUS), is based on the observation that once the training reaches a certain point, the accuracy often gets very minimal improvements from updating the weights (as the weights themselves show minuscule changes). During such periods, the accuracy can be improved by relying on updating the bias alone, while skipping weight updates. Thus, WUS effectively reduces the training time as it does not perform the gradient calculations needed for weight updates during such periods. Updating bias alone may increase the accuracy for a while, but eventually updating bias alone becomes no longer sufficient to improve the accuracy. At that point, WUS falls back to normal training in which both weights and biases are updated. This approach effectively defines two distinct phases for training process: i) normal training phase (both weights and biases are updated), and ii) WUS (only biases are updated). During the training, we may switch between these two phases based on the accuracy improvement we observe over time. 

For effectively reducing training time without impacting inversely the accuracy, we have to figure out i) when we should switch from normal training phase to WUS phase for the first time; ii) when we should switch between WUS phase and normal training phase to keep training moving (i.e., improve the accuracy); and iii) which biases should be updated (and which weights to be kept the same) during the WUS phase. Each of these decisions is a function of dataset and the NN model being trained, thus there is no static decision to be made. We should monitor the accuracy improvement over time and check when it starts to lose its momentum (which is generally characterized by minuscule weight updates).  

\subsection{Identifying Initial Epoch to Switch to WUS}
We use the standard deviation of validation accuracy to decide when to switch from normal training phase to WUS phase. We monitor the validation accuracy 
and compute the standard deviation of accuracy. 
When two successive epochs exhibit standard deviations of the validation accuracy less than some predetermined threshold (in our evaluations we use a value of 0.71\footnote{The threshold of 0.71 is empirically obtained over different models and datasets.} for this threshold), we choose the next epoch as an initial epoch for switching from normal training phase to WUS phase. 

\subsection{Switching Between Phases}
As a variant of the WUS approach, we also propose another method based on learning rate scheduler to decide when to switch back to normal training phase. In short, we call it Weight Update Skipping with Learning Rate Scheduler ($WUS+LR$). In the following subsections, we provide a detailed discussion on both $WUS$ and $WUS+LR$.

\subsubsection{Weight Update Skipping (WUS)}

\vspace{-0.2cm}
\begin{figure}[t!]
\centering
\includegraphics[height=3.5cm]{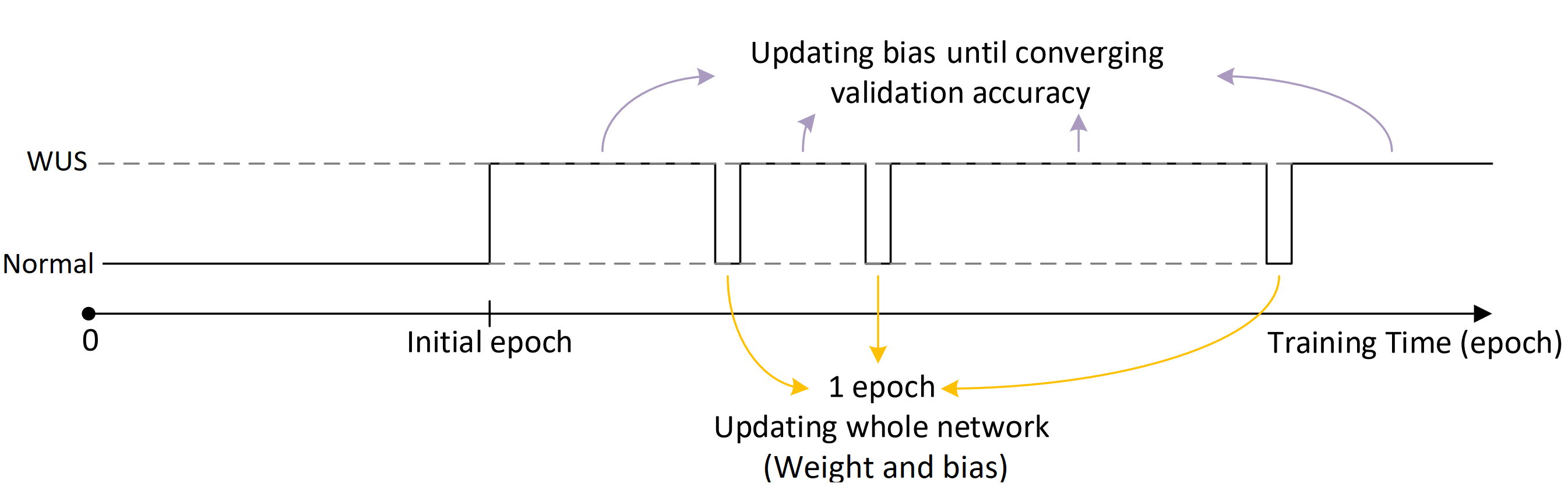}
\caption{Illustration of Switching Between Two Training Phases.}
\label{wus-workflow}
\end{figure}

After switching from normal training phase to WUS phase for the first time, we still need to keep monitoring how accuracy improves over time. When updating bias alone (in WUS phase) is no longer sufficient to keep the network improving its accuracy, we need to switch back to normal training phase. Therefore, in basic WUS variant, if the best validation accuracy does not improve after a certain number of epochs (we empirically set this number to 7), we switch back to normal training phase for only one epoch to update weight and bias for the whole network. Algorithm 1 shows the how decisions being made, and Fig.~\ref{wus-workflow} demonstrates the switching process between phases. 

\underline{\textbf{Algorithm 1}}\vspace{-0.2cm}
\begin{verbatim}
training():
    //initially WUS = False (i.e., normal training phase)
    ...
    // switch from WUS to Normal
    if (WUS == True) and (epoch > initial_epoch):
        bias/weight.requires_grad = True
        if (epoch - previous_epoch > 1) :
            WUS = False
    //switch from Normal to WUS
    if (WUS == False) and (epoch > initial_epoch):
        bias/weight.requires_grad = False
        bias/weight.grad = None
    end.
    ...
    
validation():
    ...
    //check if we need to switch between WUS and normal training phases
    WUS = WeightUpdateSkipping(7, validation_accuracy, 0, epoch)
    ...
    
WeightUpdateSkipping(patience, accuracy, delta, epoch):
  {Assuming patience is equal to 7 and delta is equal to zero};
    if best_accuracy is None:
        best_accuracy = accuracy
    else if accuracy < best_accuracy + delta:
        counter += 1
        if counter >= patience:
            WUS = True
            counter = 0
            previous_epoch = epoch
    else:
        best_accuracy = accuracy
        counter = 0
    end.
    
    return WUS
    
\end{verbatim}


\subsubsection{Weight Update Skipping with Learning Rate Scheduler (WUS+LR)}
This variant of WUS allows us to switch between phases based on the learning rate during training. Recent works advocate to use a non-monotonic learning rate instead of a constant value, claiming that using non-monotonic learning rate accelerates the convergence. Therefore, different ideas have been proposed to dynamically change the learning rate during training. In WUS+LR, we switch from WUS phase to normal training phase when the learning rate is changed. 

\subsection{Choosing Layers To Apply WUS}

\begin{figure}[htp]
\centering

\begin{subfigure}{0.49\columnwidth}
\centering
\includegraphics[width=\textwidth]{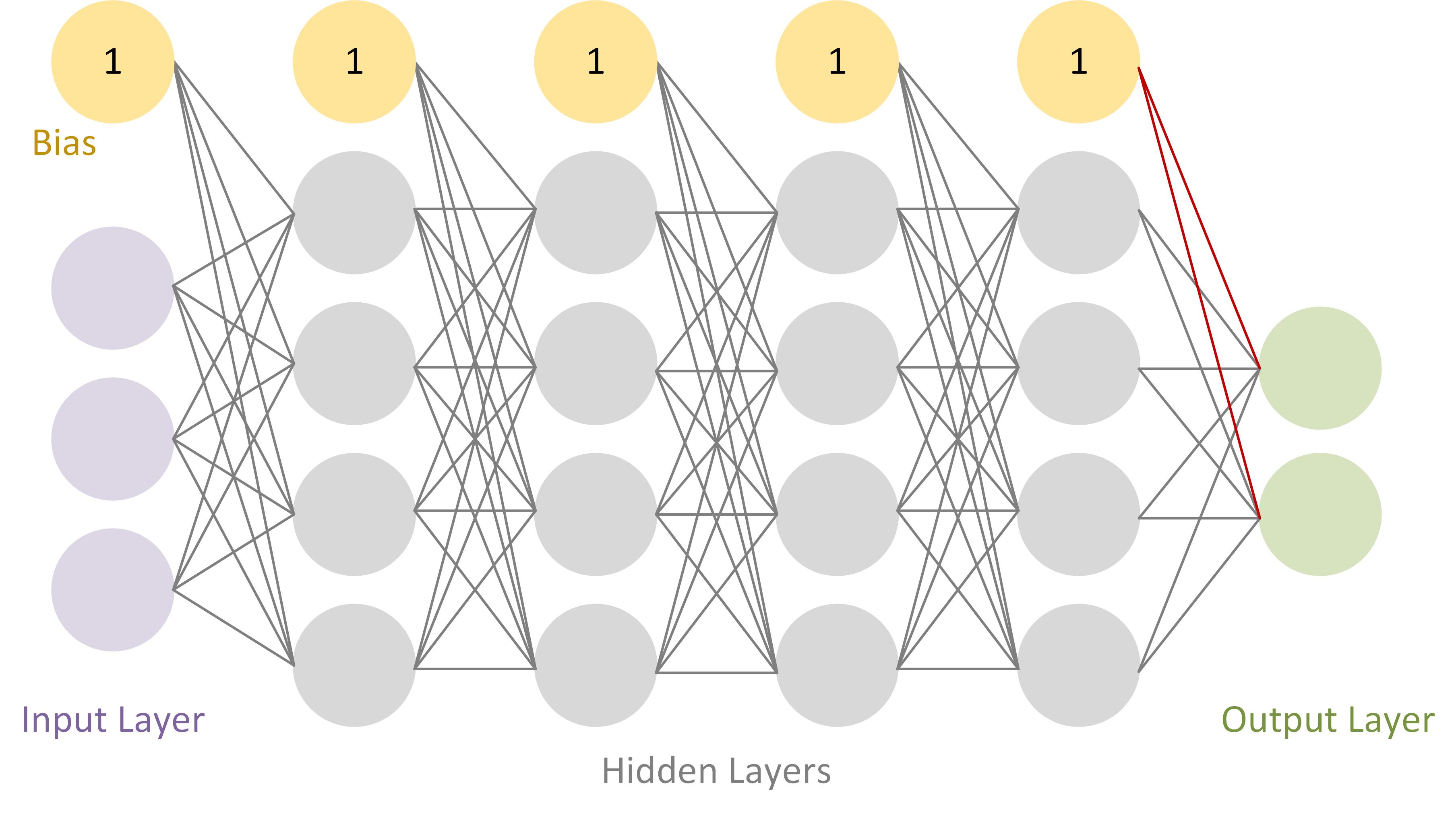}
\caption{}
\label{nn-layers-wus:l1}
\end{subfigure}\hfill
\begin{subfigure}{0.49\columnwidth}
\centering
\includegraphics[width=\textwidth]{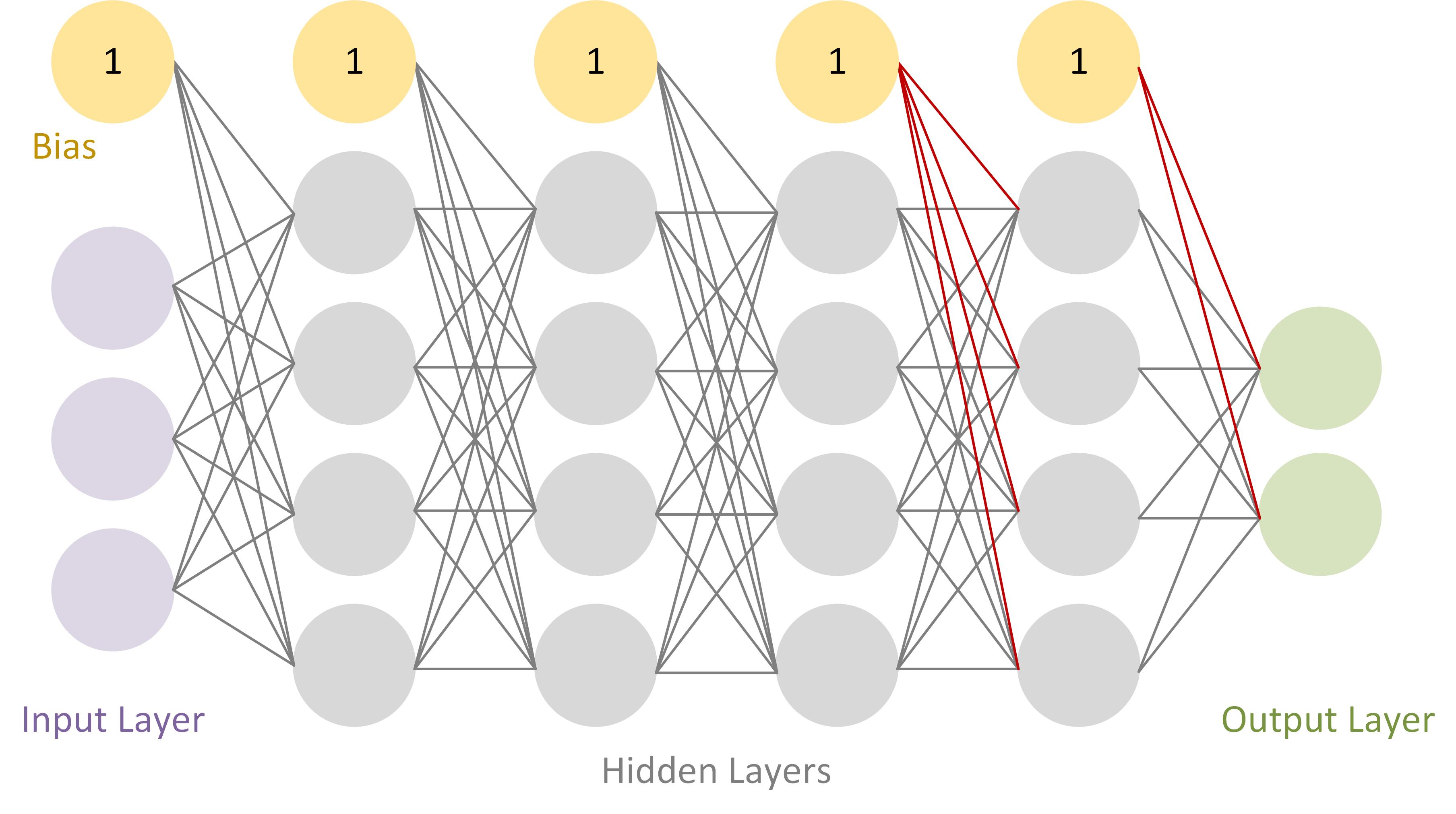}
\caption{}
\label{nn-layers-wus:l2}
\end{subfigure}

\medskip

\begin{subfigure}{0.49\columnwidth}
\centering
\includegraphics[width=\textwidth]{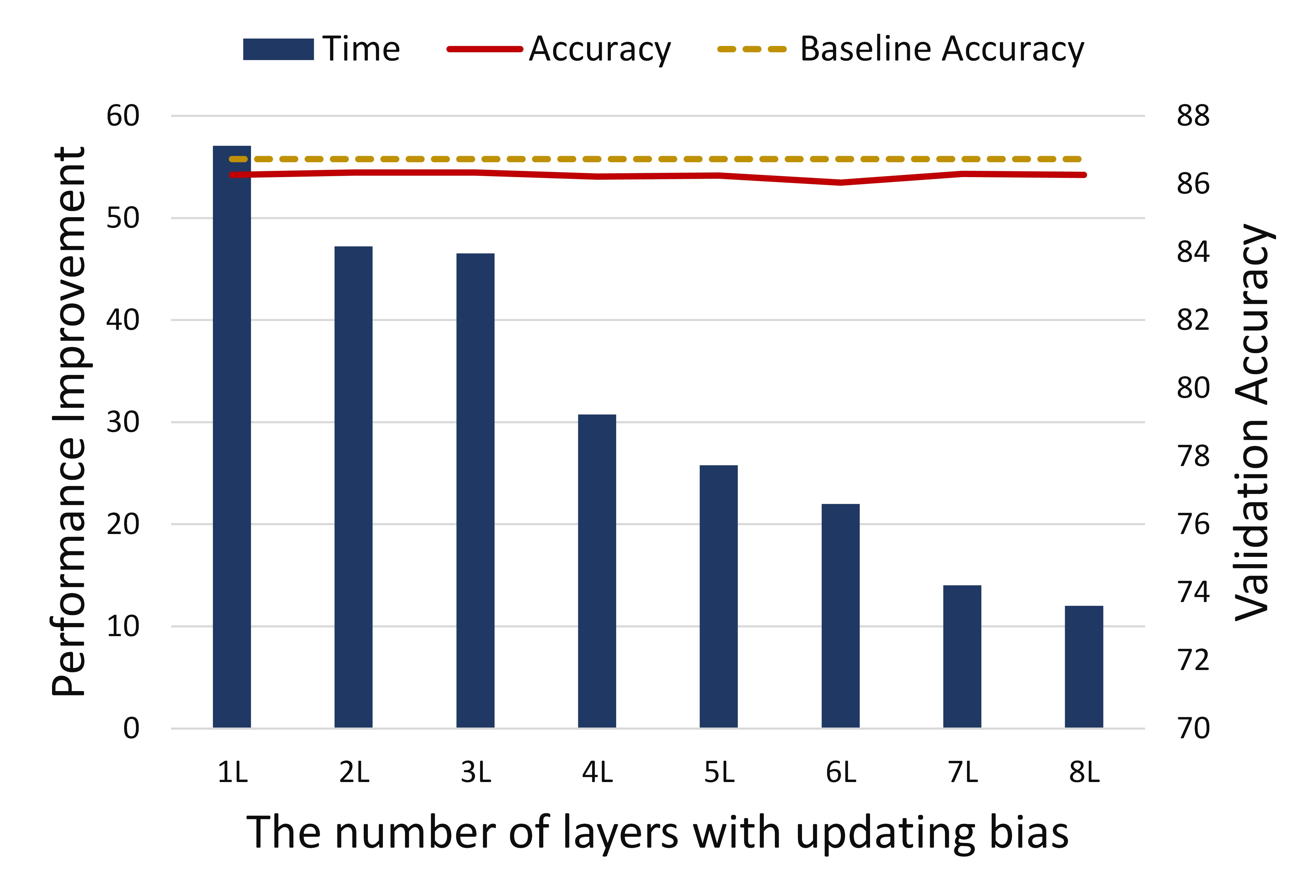}
\caption{}
\label{nn-layers-wus:alexnet}
\end{subfigure}\hfill
\begin{subfigure}{0.49\columnwidth}
\centering
\includegraphics[width=\textwidth]{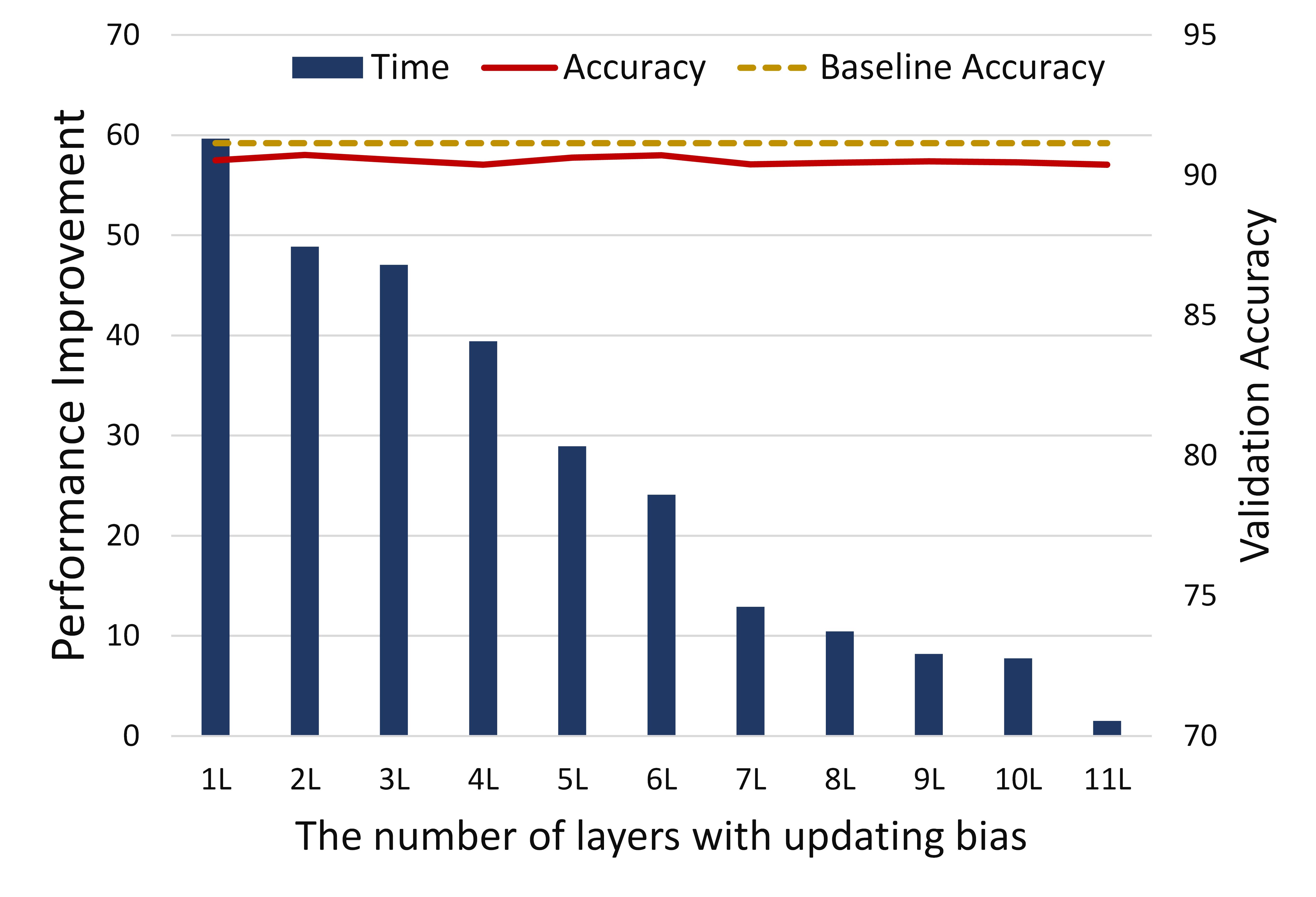}
\caption{}
\label{nn-layers-wus:vgg11}
\end{subfigure}

\medskip

\begin{subfigure}{0.49\columnwidth}
\centering
\includegraphics[width=\textwidth]{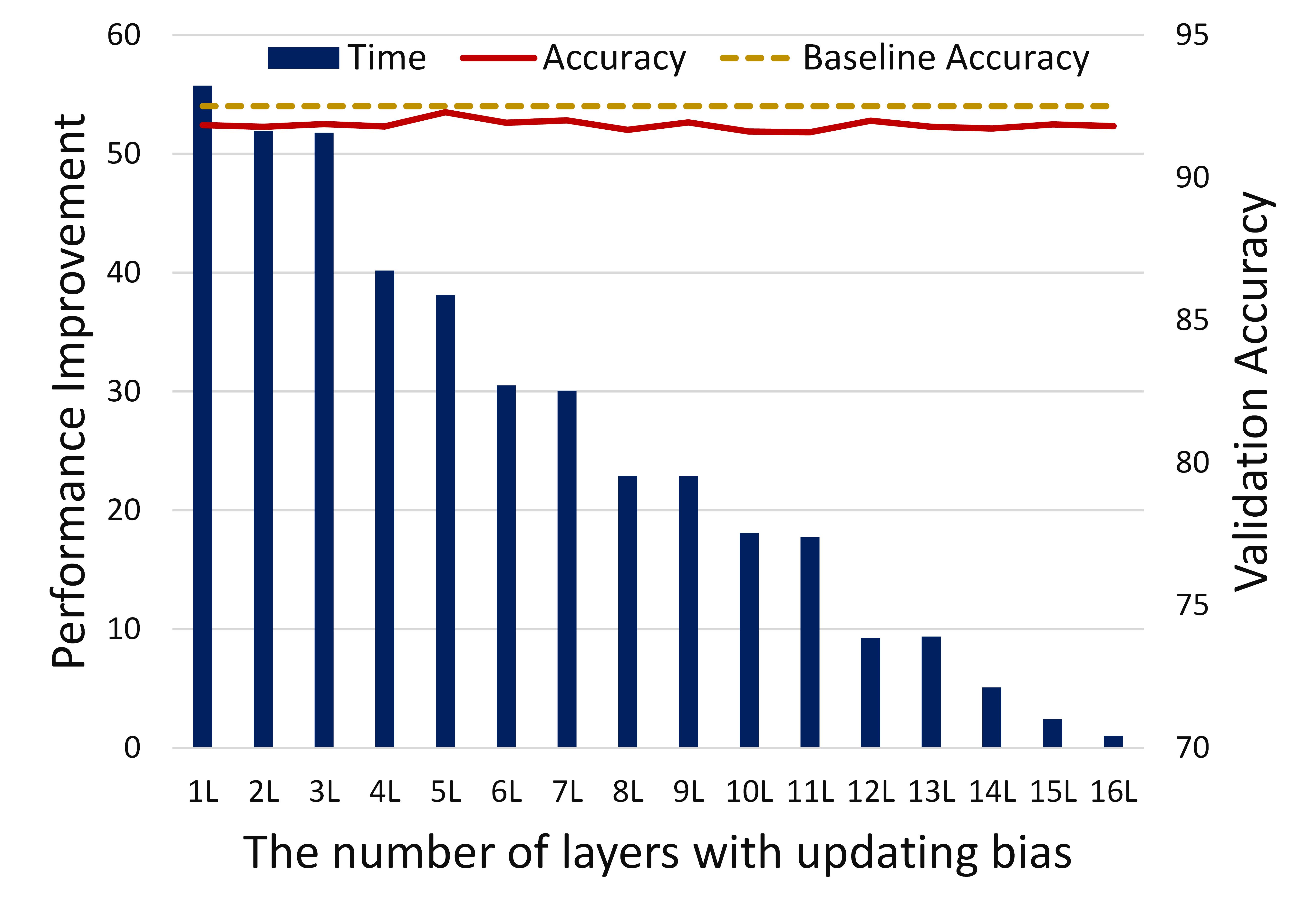}
\caption{}
\label{nn-layers-wus:vgg16}
\end{subfigure}

\caption{Illustration of (a) skipping weight and bias updates, except last layer in which only the biases are updated (i.e., 1L); (b) skipping weight and bias updates, except the last two layers in which only the biases are updated (i.e., 2L). Training time reduction and accuracy change as we change the numbers of layers in which weight and bias updates are skipped in WUS phase for (c) ALexNet, (d) VGG-11, and (e) VGG-16, with CIFAR-10.}
\label{nn-layers-wus}

\end{figure}

After identifying the initial epoch for switching to WUS phase, we may need to decide whether we need to apply WUS to all layers or some subset of the layers. The obvious tradeoff in this decision is between the accuracy and training time. Intuitively, the training time should reduce more as we skip more layers and their weight updates. On the other hand, the accuracy may not improve enough if there is not sufficient number of weights being updated. To determine the impact of the number of layers subject to WUS, we performed the following analysis. Starting from the last hidden layer, we repeated the analysis and apply WUS to more layers in an incremental fashion, and monitor both training time and accuracy changes. As an example, for the first analysis, we update the biases only in the last layer during the backward pass of the training process (assuming we switch to WUS phase), while skipping weight and bias updates for the all layers (down to the very first layer). This is illustrated in Fig.~\ref{nn-layers-wus:l1}, where the red lines indicate biases of the last layer that are updated in WUS phase.  Then, we repeated the same analysis with updating the biases of the last two layers, while skipping all the weight and bias updates for the remaining of the layers (again, down to the very first layer) in WUS phase. This is illustrated in Fig.~\ref{nn-layers-wus:l2} (similar to Fig.~\ref{nn-layers-wus:l1}, red lines indicate biases that are updated; in this case, the biases of the last two layers). We repeated these analysis until we reach to the very first layer of the models we used for this paper (i.e., AlexNet, VGG-11, and VGG-16). To differentiate each of these analysis, we use the following labeling: 1L, 2L, 3L, ... where 1L indicates the analysis in which only the last layer's biases are updated; 2L indicates the analysis in which only the last two layers' biases are updated during WUS phase, and so on.  Corresponding training time reduction (and validation accuracy) results are shown in Figs.~\ref{nn-layers-wus:alexnet},~\ref{nn-layers-wus:vgg11}, and ~\ref{nn-layers-wus:vgg16} for AlexNet, VGG-11, and VGG-16, respectively~\footnote{each experiment is repeated 15 times, and the averages are reported here. CIFAR-10 dataset is used.}. It is evident that as we have more layers in which we skip weight updates, higher training time reduction can be achieved (more than 50\% reduction in training time for 1L), while the accuracy remains less sensitive to the number of layers subject to weight update skipping. This observation motivates to skip weight updates in more layers during WUS phase (as the accuracy seems to be less sensitive) for reducing the training time to the extend possible (as in the case of 1L).

\section{Evaluation}
\subsection{Implementation Details}
We empirically evaluated the effectiveness of proposed Weight Update Skipping (WUS) optimization on some of the state-the-art ML models (i.e., ResNet-18, VGG-16, VGG-11, and AlexNet) trained for two datasets: CIFAR-10, CIFAR-100~\cite{krizhevsky2009learning}. CIFAR-10 dataset is made up of 60,000 32-by-32 color images that are categorized into 10 classes. The training set and validation set consist of 50,000 and 10,000 images, respectively. CIFAR-100 dataset is similar to CIFAR-10, except it is categorized into 100 classes. For both datasets, we use the same experimental settings. We train for both datasets without data augmentation. We use batch normalization for the ResNet-18 model. We used x86\_64 architecture (Intel(R) Xeon(R) CPU E5-2680 v4 2.40GHz) with 10 cores and 3 level caches.
We use PyTorch  Framework for our evaluations. All baseline models (i.e., w/o WUS) are trained with Stochastic Gradient Descent (SGD) with a mini-batch size of 128 for 200 epochs. We use the initial learning rate of 0.1 and divide it by a factor of 10 after 30 epochs. The weight decay is $1e^{-4}$ and momentum is 0.9 in our evaluations. For WUS and WUS+LR evaluations, we keep the structure of networks and all configuration parameters same as the baseline. All evaluations are repeated 15 times and we reported the average of them (both for training time and accuracy).

\subsection{Training Time}
We compare the efficiency WUS and WUS+LR to baseline. Table~\ref{table:time} presents training time (based on the second) of two proposed techniques and baseline for both CIFAR-10 and CIFAR-100 datasets. These results are coming from a setting when a model employs WUS and WUS+LR in the last layer.
On average, our proposed methods WUS and WUS+LR reduced the training time (compared to the baseline)  by 54\%, and 50\%, respectively on CIFAR-10; and 43\% and 35\% on CIFAR-100, respectively. 

\setlength{\tabcolsep}{4pt}
\begin{table}
\begin{center}
\caption{Training time comparison for proposed training approaches on different datasets}
\label{table:time}
\begin{tabular}{llll}
\hline\noalign{\smallskip}
Model & CIFAR-10 & CIFAR-100 \\
     & time (\% reduction) & time (\% reduction) \\
\noalign{\smallskip}
\hline
\noalign{\smallskip} 
VGG-11 (Baseline)  & 50239.42 & 54382.03 \\
VGG-11 (WUS)  &  20632.78 (\%58.93) &  27376.36 (\%49.66) \\
VGG-11 (WUS+LR)  & 24303.06 (\%51.6) & 30826.08 (\%43.32) \\
\noalign{\smallskip}
\hline
\noalign{\smallskip}
VGG-16 (Baseline)  & 71624.71 & 77125.61 \\
VGG-16 (WUS)  &  35777.20 (\%50.05) & 48480.81 (\%37.14) \\
VGG-16 (WUS+LR)  & 38238.55 (\%46.61) & 50864.71 (\%34.05) \\
\noalign{\smallskip}
\hline
\noalign{\smallskip}
AlexNet (Baseline) & 27978.22 & 26496.81\\ 	
AlexNet (WUS) & 11609.07 (\%58.5) & 15249.07 (\%42.45)\\
AlexNet (WUS+LR) & 12768.27 (\%54.36) & 15894.22 (\%40.01 )\\
\noalign{\smallskip}
\hline
\noalign{\smallskip}
ResNet-18 (Baseline) & 114167.84  & 124305.02 \\
ResNet-18 (WUS) & 57665.51 (\%49.49) & 71163.48 (\%42.75)\\
ResNet-18 (WUS+LR) & 59131.44 (\%48.21) & 94423.88 (\%24.04)\\
\hline
\end{tabular}
\end{center}
\end{table}
\setlength{\tabcolsep}{1.4pt}

\subsection{Accuracy}
The accuracy results for WUS and WUS+LR are shown in Table~\ref{table:acc}.
Similar to training time, these results are coming from a setting when a model employs WUS and WUS+LR in the last layer.  As can be seen, WUS and WUS+LR obtain almost the same accuracy as the baseline with less than 1\% drop in accuracy on average. One may notice that WUS+LR provides slightly better accuracy than WUS (0.18\%); although WUS reduces the training time more than WUS+LR (by 6\%).
The fact that WUS+LR provides better accuracy compared to WUS can be explained by its ability to capture the stagnation of accuracy earlier than WUS and reacts to it by switching to normal training in a timely manner. However, this ability comes with its slightly lower performance (i.e., higher training time) compared to WUS, as it causes training to spend more time on normal training phase (thus, more weight updates).

\begin{figure}
  \begin{minipage}[b]{0.5\linewidth}
    \centering
    \subcaptionbox{VGG-11 \label{cifar10-accuracy:vgg11} }
    {\includegraphics[width=\linewidth]{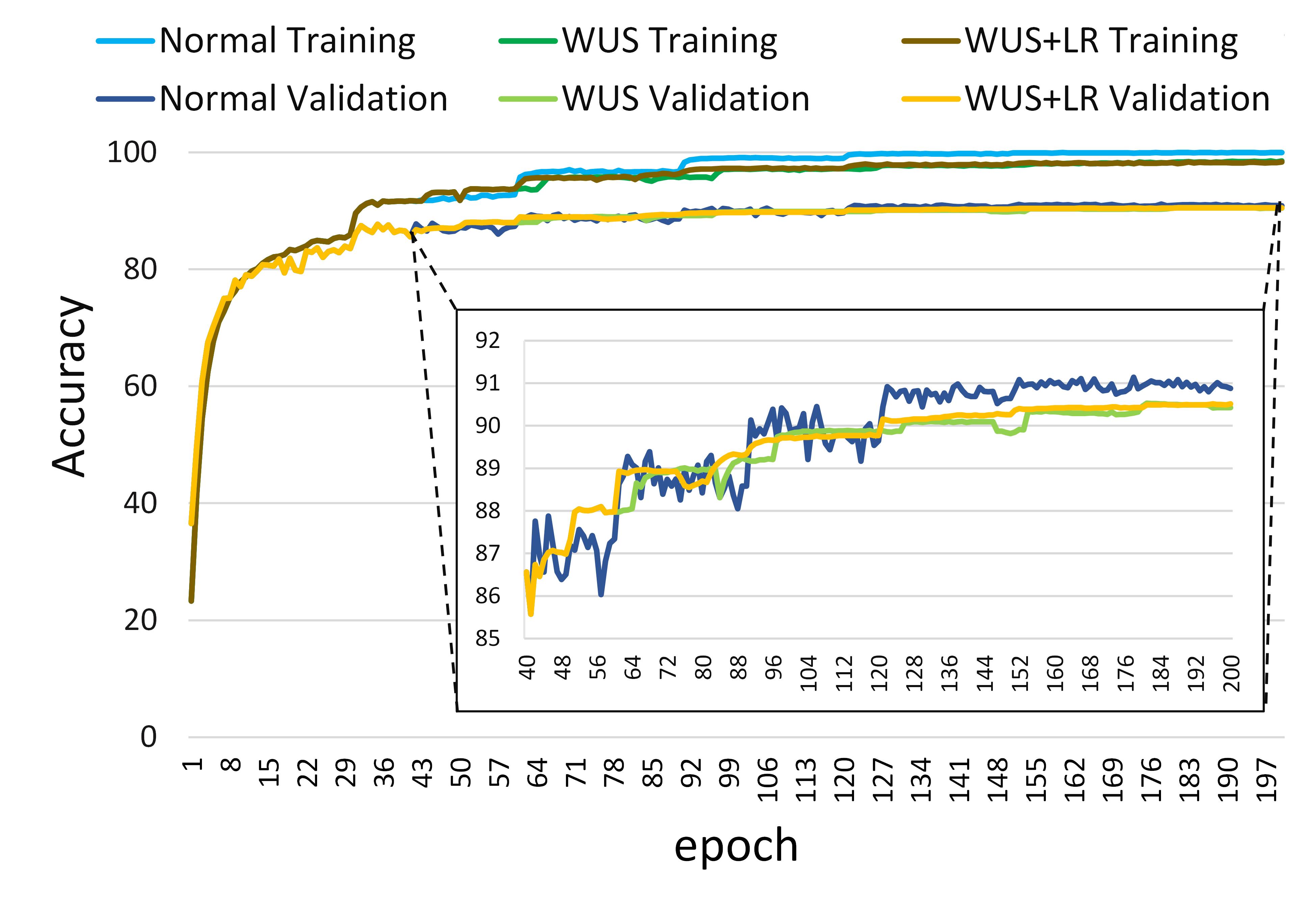}}
    
  \end{minipage}
  \begin{minipage}[b]{0.5\linewidth}
    \centering
    \subcaptionbox{VGG-16 \label{cifar10-accuracy:vgg16} }
    {\includegraphics[width=\linewidth]{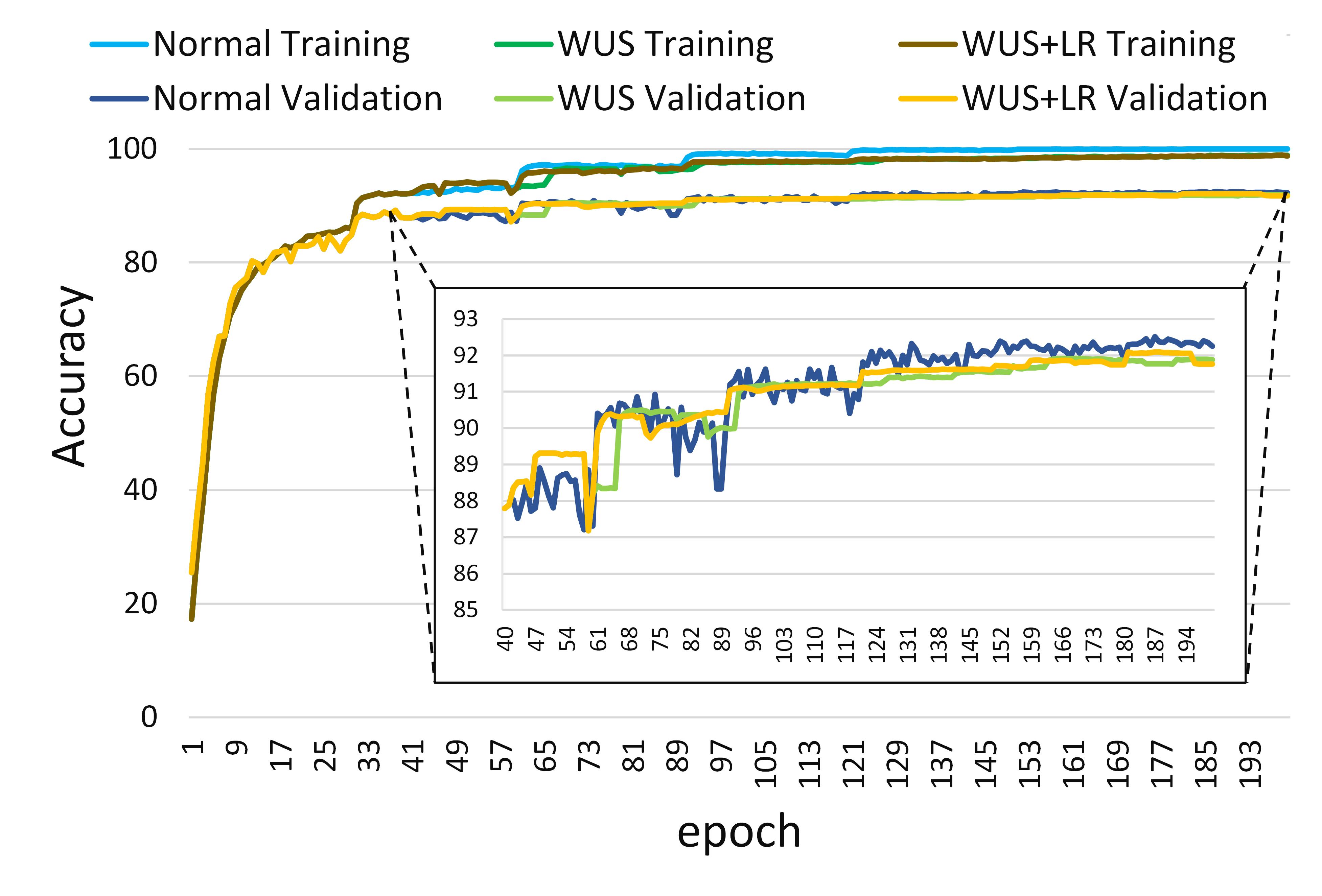}}
    
  \end{minipage} 
  \begin{minipage}[b]{0.5\linewidth}
    \centering
    \subcaptionbox{AlexNet \label{cifar10-accuracy:alexnet}} 
    {\includegraphics[width=\linewidth]{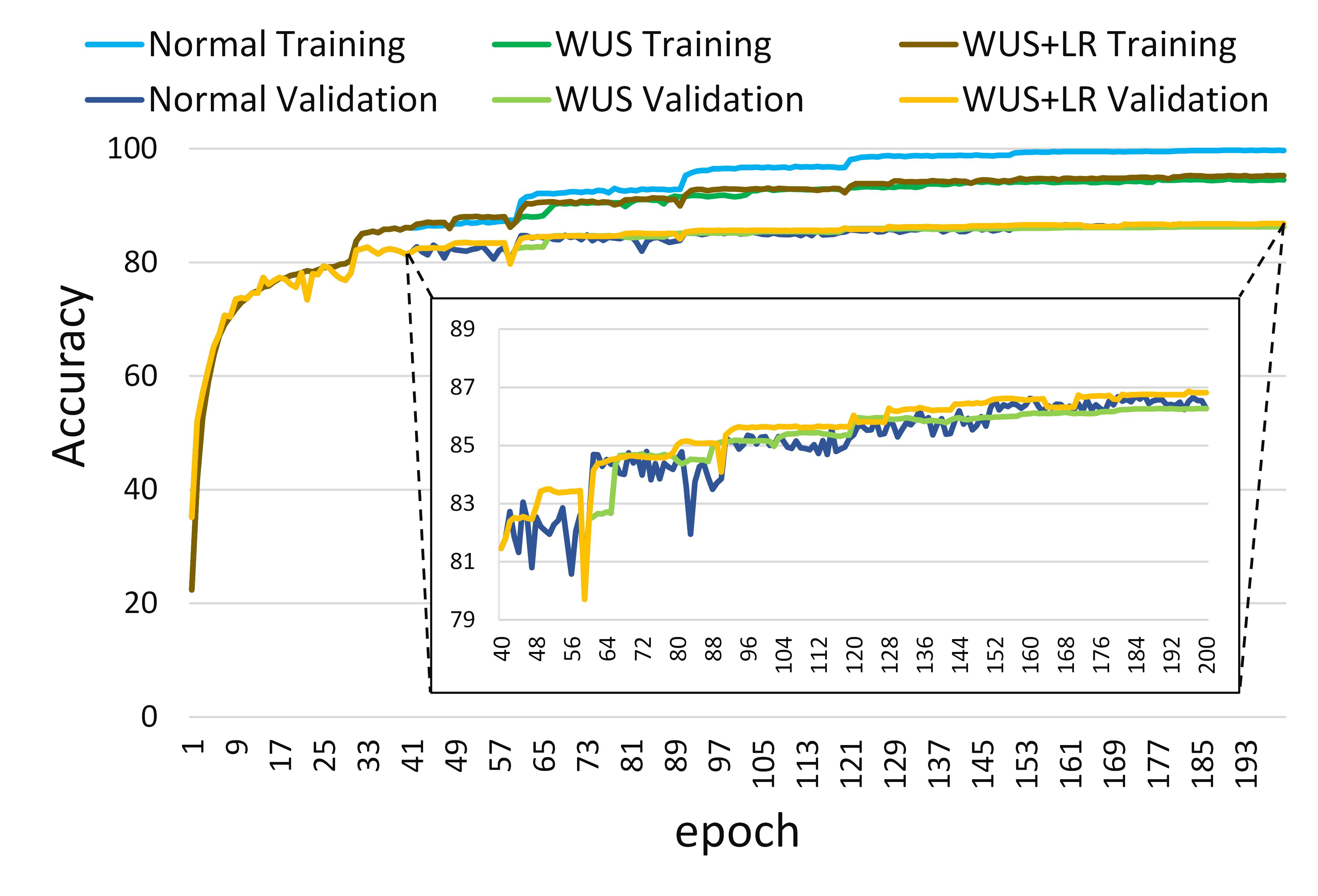}}
    
  \end{minipage}
  \begin{minipage}[b]{0.5\linewidth}
    \centering
    \subcaptionbox{ResNet-18 \label{cifar10-accuracy:resnet}} 
    {\includegraphics[width=\linewidth]{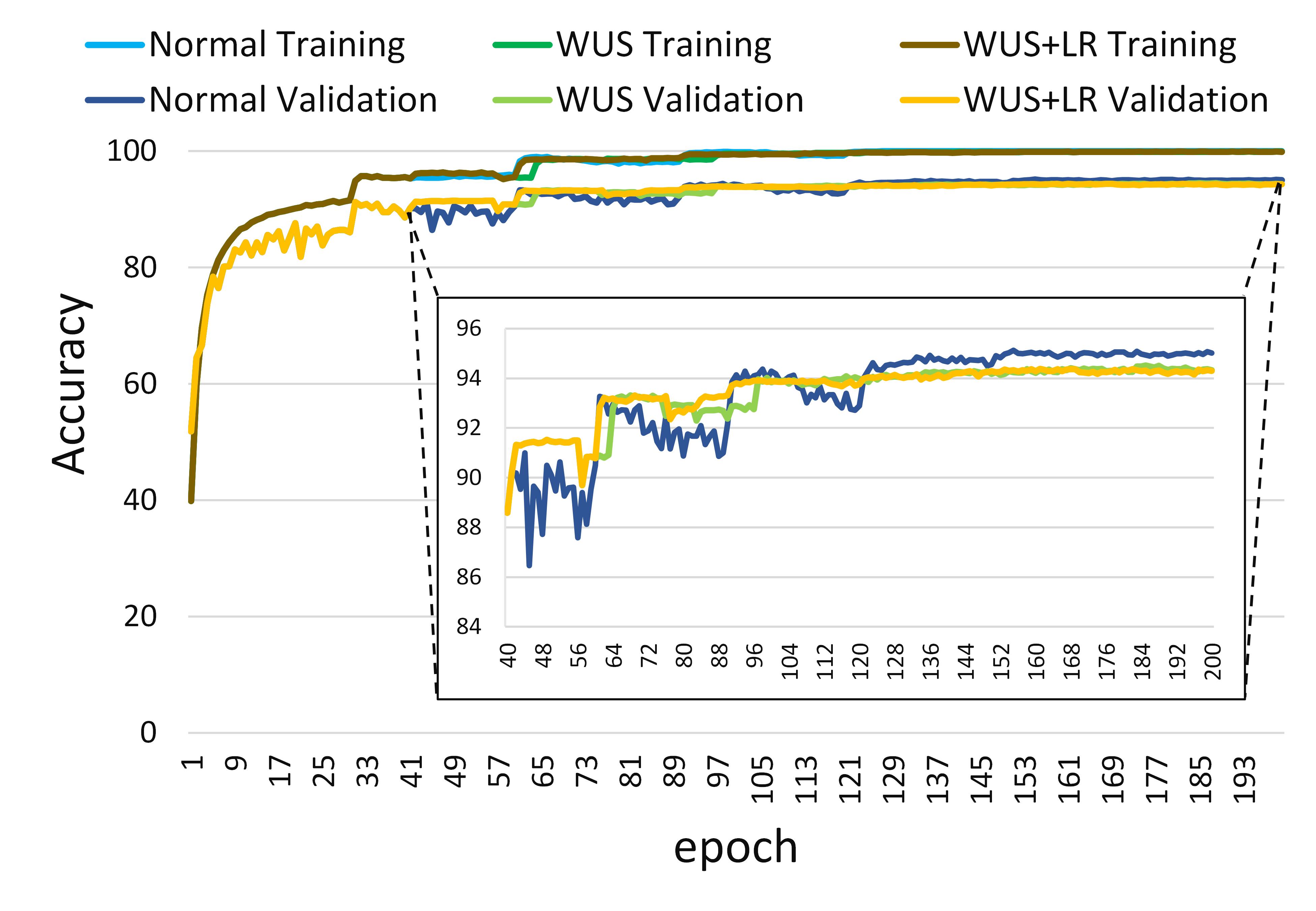}}
    
  \end{minipage} 
  \caption{Validation and training accuracy on VGG-11, VGG-16, AlexNet, and ResNet-18 for CIFAR-10 dataset.}
  \label{cifar10-accuracy} 
\end{figure}

\begin{figure}
  \begin{minipage}[b]{0.5\linewidth}
    \centering
    \subcaptionbox{VGG-11 \label{cifar100-accuracy:vgg11}}
    {\includegraphics[width=\linewidth]{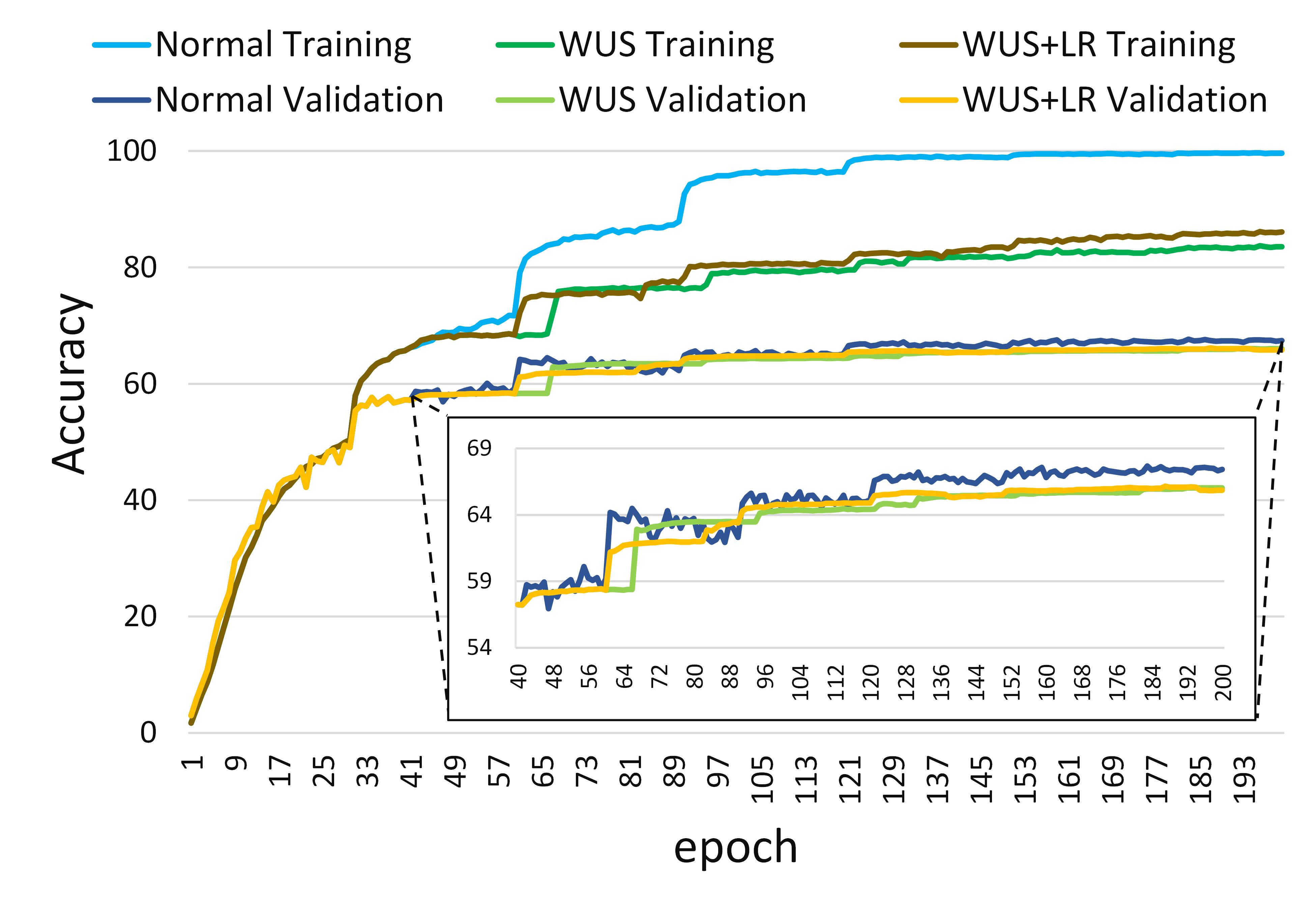}}
  \end{minipage}
  \begin{minipage}[b]{0.5\linewidth}
    \centering
    \subcaptionbox{VGG-16 \label{cifar100-accuracy:vgg16} }
    {\includegraphics[width=\linewidth]{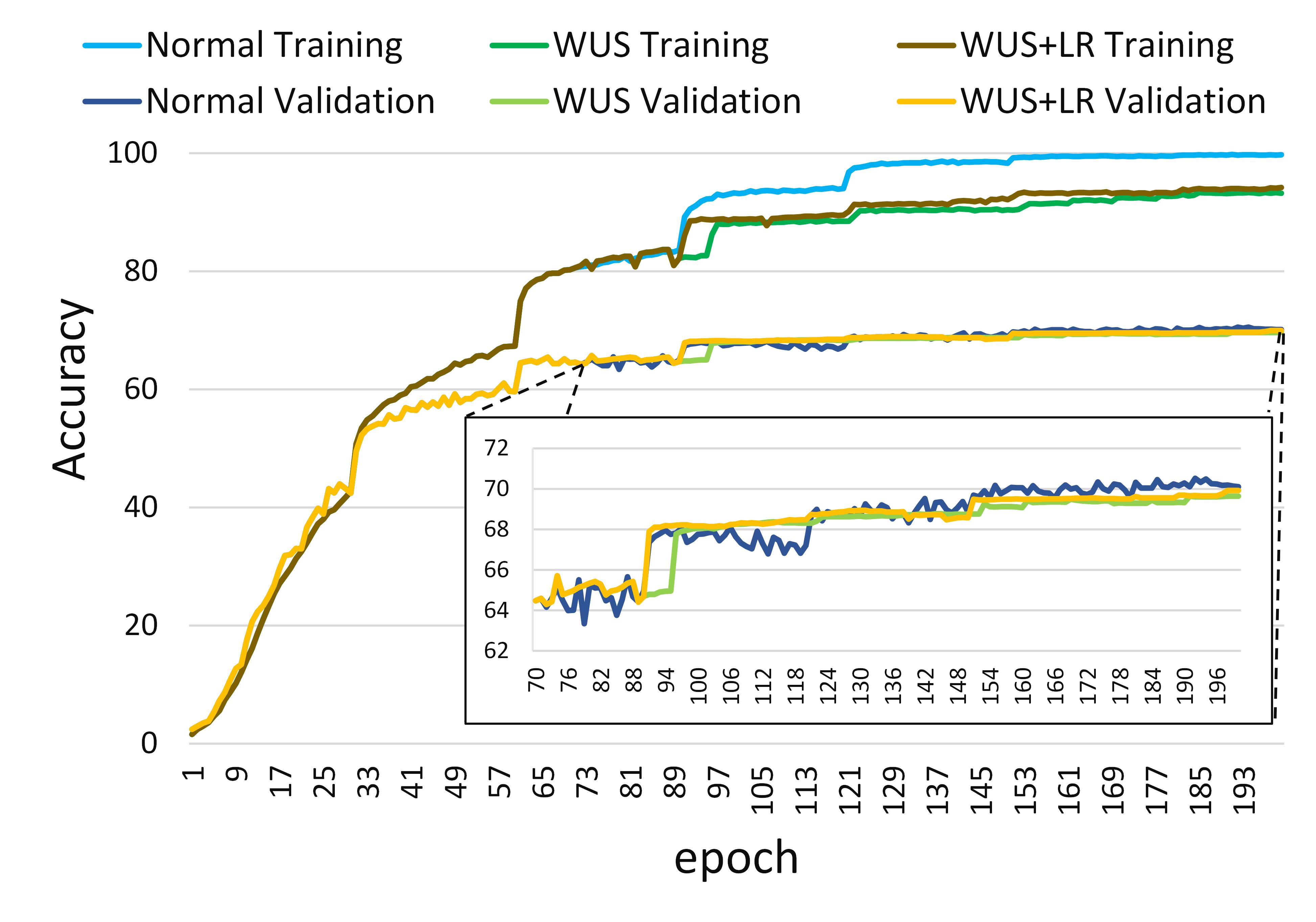}}
    
  \end{minipage} 
  \begin{minipage}[b]{0.5\linewidth}
    \centering
     \subcaptionbox{AlexNet \label{cifar100-accuracy:alexnet} }
    {\includegraphics[width=\linewidth]{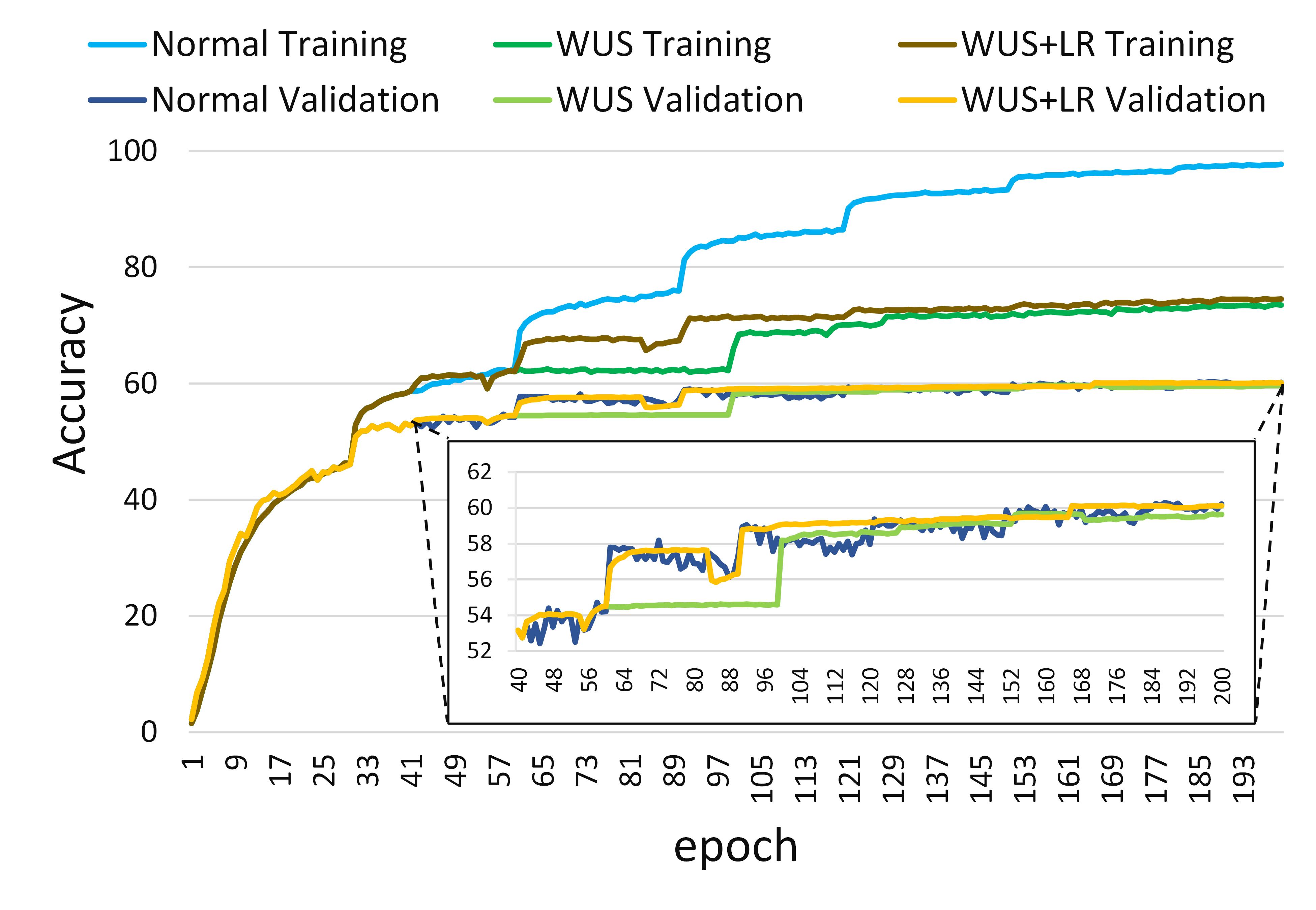}}
   
  \end{minipage}
  \begin{minipage}[b]{0.5\linewidth}
    \centering
    \subcaptionbox{ResNet-18 \label{cifar100-accuracy:resnet} }
    {\includegraphics[width=\linewidth]{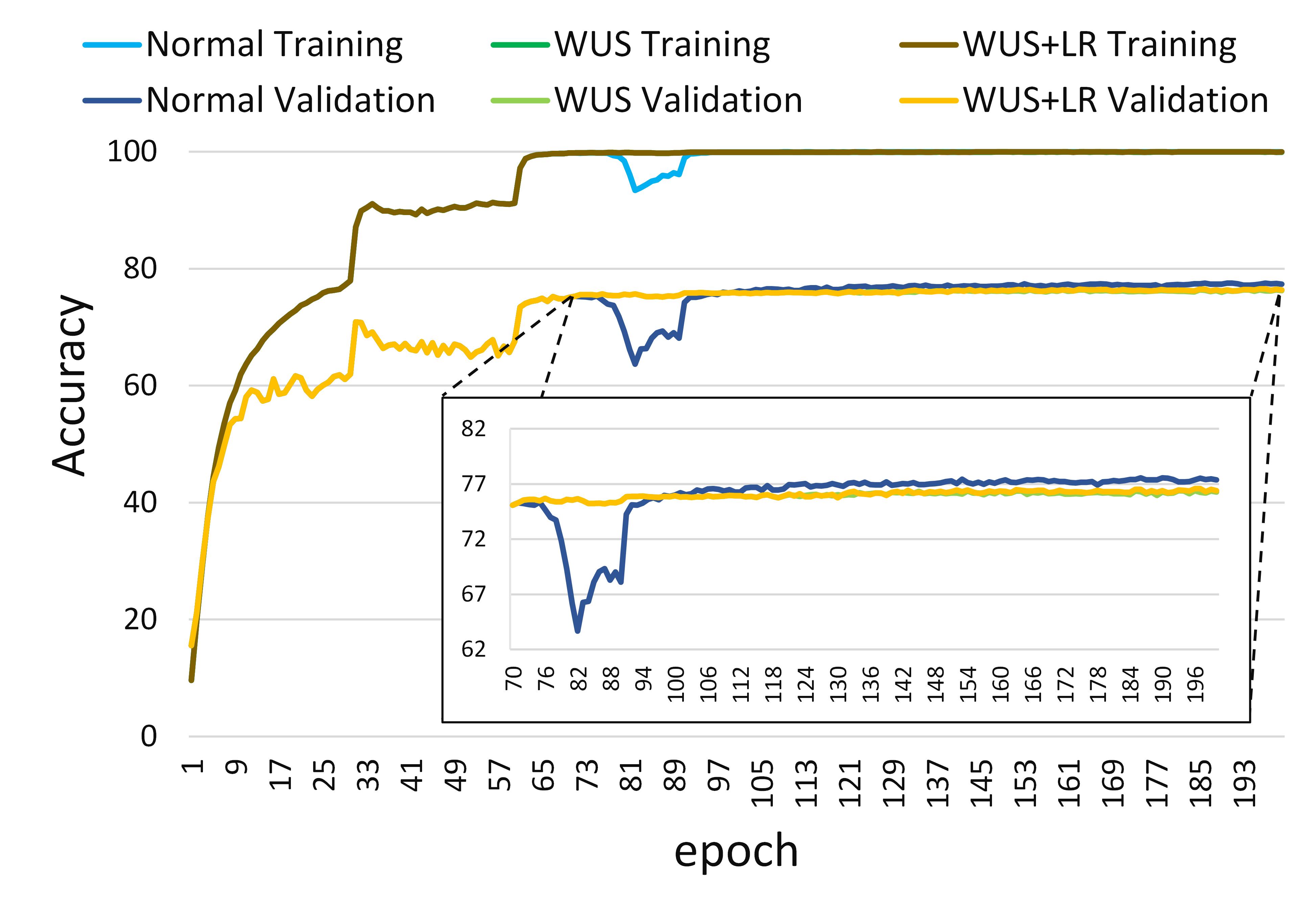}}
    
  \end{minipage} 
  \caption{Validation and training accuracy on VGG-11, VGG-16, AlexNet, and ResNet-18 for CIFAR-100 dataset.}
  \label{cifar100-accuracy} 
\end{figure}

Fig.~\ref{cifar10-accuracy} shows how validation accuracy and training accuracy of four models for CIFAR-10 dataset evolve over time when WUS and WUS+LR are employed. As can be seen for the AlexNet model, WUS, and WUS+LR methods avoid over-fitting by decreasing the gap between training and validation accuracy. Furthermore, since these two methods have less fluctuation compared to baseline accuracy, they can converge faster than baseline training. Similarly, Fig.~\ref{cifar100-accuracy}  shows how validation accuracy and training accuracy of four models for CIFAR-100 dataset evolve over time when WUS and WUS+LR are employed. The very same trends are obsersed as in the case of CIFAR-10 dataset.

\setlength{\tabcolsep}{4pt}
\begin{table}
\begin{center}
\caption{Validation Accuracy comparison on benchmark datasets}
\label{table:acc}
\begin{tabular}{llll}
\hline\noalign{\smallskip}
Model & CIFAR-10 & CIFAR-100 \\
\noalign{\smallskip}
\hline
\noalign{\smallskip}
VGG-11 (Baseline)  & 91.30 & 67.85 \\
VGG-11 (WUS)  & 90.63 (\%0.67) & 66.00 (\%1.85) \\
VGG-11 (WUS+LR)  & 90.75 (\%0.55) & 66.36 (\%1.49)\\
\noalign{\smallskip}
\hline
\noalign{\smallskip}
VGG-16 (Baseline)  & 92.68 & 70.82 \\
VGG-16 (WUS)  & 91.84 (\%0.84) & 69.6 (\%1.22) \\
VGG-16 (WUS+LR)  & 92.00 (\%0.68) & 69.79 (\%1.03) \\
\noalign{\smallskip}
\hline
\noalign{\smallskip}
AlexNet (Baseline) & 86.86 & 60.27 \\
AlexNet (WUS) & 86.27 (\%0.59) & 59.97 (\%0.3)\\
AlexNet (WUS+LR) & 86.39 (\%0.47) & 60.20 (\%0.07)\\
\noalign{\smallskip}
\hline
\noalign{\smallskip}
ResNet-18 (Baseline) & 95.25 & 77.39\\
ResNet-18 (WUS) & 94.51(\%0.74) & 76.56 (\%0.83)\\
ResNet-18 (WUS+LR) & 94.58 (\%0.67) & 76.73 (\%0.66)\\
\hline
\end{tabular}
\end{center}
\end{table}
\setlength{\tabcolsep}{1.4pt}

\subsection{Training Parameters}
This section compares the number of parameters that are updated when using different training approaches for different network models and datasets. 


The total number of trainable parameters, say (${N}$) can be calculated by adding up the parameters updated in normal training phase and in WUS phase. This is shown in equation~\ref{equation-training-params} where ${E_{wus}}$, and ${E_{normal}}$ are the number of epochs that WUS phase (i.e., only biases of the last x layers are updated; where x is 1 for 1L; x is 2 for 2L, and so on), and normal training phase (i.e., both weights and biases of the all layers are updated) are employed, respectively. Likewise, ${w}$, and $b$ is the number of weights and biases exist in the network. 

\begin{equation}
    N = (E_{wus} * b) + (E_{normal} * (b + w))   
    \label{equation-training-params}
\end{equation} 

Table~\ref{table:eng} shows the percentage of reduction in updated parameters for the proposed WUS and WUS+LR compare to baseline training on CIFAR datasets. The number of updated parameters is reduced 72\% on ResNet-18, and 74\% on other networks when WUS is employed; and reduced 71\% on VGG-16 and ResNet-18, 72\% on AlexNet, and 73\% on VGG-11 when WUS+LR is employed for CIFAR-10 dataset.  Likewise, the number of updated parameters is reduced 56\% on AlexNet, 60\% on VGG-16, 65\% on ResNet-18, and 76\% on VGG-11 when WUS is employed; and reduced 54\% on ResNet-18, 56\% on VGG-16, 73\% on VGG-11, and 74\% on AlexNet when WUS+LR is employed on CIFAR-100 dataset.

The considerable amount of reduction on number of updated parameters would likely to yield higher energy efficiency (along with the reduced training time) which would allow to fit larger neural networks within a given power budget, that would eventually either increase accuracy or help to tackle more complex problems. We keep energy efficiency analysis for the proposed training methodology as a future work.

\setlength{\tabcolsep}{4pt}
\begin{table}
\begin{center}
\caption{The percentage reduction of updated parameters comparison on CIFAR datasets.}
\label{table:eng}
\begin{tabular}{lllllllll}
\hline\noalign{\smallskip}
Model & \multicolumn{2}{c}{CIFAR-10} & \multicolumn{2}{c}{CIFAR-100}\\
\cline{2-5}
 & WUS & WUS+LR & WUS & WUS+LR\\
\hline
\noalign{\smallskip}
VGG-11 &  \%74 & \%73 & \%76 & \%73 \\

VGG-16  & \%74 & \%71 & \%60 & \%56 \\

AlexNet & \%74 & \%72 & \%56 & \%74 \\

ResNet-18 & \%72 & \%71 & \%65 & \%54\\
\hline
\end{tabular}
\end{center}
\end{table}
\vspace{-0.2cm}

\section{Conclusions}
We propose a new training methodology called Weight Update Skipping (WUS) based on the observation that the improvement of accuracy shows temporal variations, thus motivating us to react this time-dependent variation accordingly by identifying which parameters should be updated. In essence,  WUS suggests to skip updating weights when the accuracy improvement is minuscule (along with corresponding gradient calculations). Although weights remain the same, the biases are kept updated, ensuring the training still makes forward progress and avoids overfitting. While reducing the training time considerably, WUS virtually keeps the accuracy intact compared to baseline training on various state-of-the-art neural network models for CIFAR datasets.

\bibliographystyle{unsrt}  
\bibliography{template}  

\end{document}